\pdfoutput=1

\documentclass[11pt]{article}

\usepackage[final]{acl}

\usepackage{times}
\usepackage{latexsym}

\usepackage{booktabs}
\usepackage{hyperref}
\usepackage{url}
\usepackage{siunitx}
\usepackage{multirow}
\usepackage{amssymb}   

\usepackage[T1]{fontenc}

\usepackage[utf8]{inputenc}

\usepackage{microtype}

\usepackage{inconsolata}

\usepackage{graphicx}
\usepackage{xcolor}
\usepackage{pifont}
\newcommand{\cmark}{\ding{51}}
\usepackage{listings}
\usepackage{fvextra}
\definecolor{codegreen}{rgb}{0,0.6,0}
\definecolor{codegray}{rgb}{0.5,0.5,0.5}
\definecolor{codepurple}{rgb}{0.58,0,0.82}
\definecolor{backcolour}{rgb}{0.95,0.95,0.92}

\lstdefinestyle{mystyle}{
    backgroundcolor=\color{backcolour},
    commentstyle=\color{codegreen},
    keywordstyle=\color{magenta},
    numberstyle=\tiny\color{codegray},
    stringstyle=\color{codepurple},
    basicstyle=\ttfamily\small, 
    breakatwhitespace=false,
    breaklines=true,
    captionpos=b,
    keepspaces=true,
    numbers=left,
    numbersep=5pt,
    showspaces=false,
    showstringspaces=false,
    showtabs=false,
    tabsize=2,
    frame=single, 
    framerule=0pt, 
    framesep=5pt, 
    rulesepcolor=\color{codegray}, 
    aboveskip=1.5em, 
    belowskip=1.5em, 
    linewidth=\textwidth, 
    xleftmargin=0pt, 
    xrightmargin=0pt 
}

\title{The Pursuit of Empathy: Evaluating Small Language Models \\ for PTSD Dialogue Support}

\author{
  Suhas BN\textsuperscript{1} \quad
  Yash Mahajan\textsuperscript{1} \quad
  Dominik Mattioli\textsuperscript{1} \quad
  Andrew M. Sherrill\textsuperscript{2} \\
  Rosa I. Arriaga\textsuperscript{3} \quad
  Chris W. Wiese\textsuperscript{4} \quad
  Saeed Abdullah\textsuperscript{1} \\
  \textsuperscript{1}College of Information Sciences and Technology, Penn State University, USA \\
  \textsuperscript{2}Department of Psychiatry and Behavioral Sciences, Emory University, USA \\
  \textsuperscript{3}School of Interactive Computing, Georgia Tech, USA \\
  \textsuperscript{4}School of Psychology, Georgia Tech, USA \\
  \texttt{\{bnsuhas,saeed\}@psu.edu},  \texttt{andrew.m.sherrill@emory.edu}
}

\begin{document}
\maketitle
\begin{abstract}
\vspace{-0.2cm}
This paper investigates the capacity of small language models (0.5B-5B parameters) to generate empathetic responses for individuals with PTSD. We introduce Trauma-Informed Dialogue for Empathy (TIDE), a novel dataset comprising 10,000 two-turn conversations across 500 diverse, clinically-grounded PTSD personas (\url{https://huggingface.co/datasets/yenopoya/TIDE}). Using frontier model outputs as ground truth, we evaluate eight small LLMs in zero-shot settings and after fine-tuning. Fine-tuning  enhances empathetic capabilities, improving cosine similarity and perceived empathy, although gains vary across emotional scenarios and smaller models exhibit a ``knowledge transfer ceiling.'' As expected, Claude Sonnet 3.5 consistently outperforms all models, but surprisingly, the smaller models often approach human-rated empathy levels. Demographic analyses showed that older adults favored responses that validated distress before offering support ($p = .004$), while graduate-educated users preferred emotionally layered replies in specific scenarios. Gender-based differences were minimal ($p > 0.15$), suggesting the feasibility of broadly empathetic model designs. This work offers insights into building resource-efficient, emotionally intelligent systems for mental health support.
\end{abstract}

\vspace{-0.1cm} 
\section{Introduction}
\vspace{-0.2cm}
Conversational AI systems are increasingly embedded in daily life but often lack a crucial quality: empathy, the ability to recognize and validate users' emotional experiences \citep{aher2023using, 6sharma2022partner}. This shortfall is especially significant in sensitive contexts like mental health support, where users managing conditions such as Post-Traumatic Stress Disorder (PTSD) seek understanding over factual assistance \citep{rabin2009cbt}. Large language models (LLMs) have shown promise in approximating empathetic communication. However, their massive scale (often >100B parameters) poses barriers to deployment in settings with privacy, latency, and compute constraints \citep{1tahir2023cbt}. In contrast, small language models (SLMs) offer practical advantages like on-device deployment and faster inference, but raise a critical question: can models with 0.5B-5B parameters deliver meaningful empathetic support?

This study investigates four research questions:

\textbf{RQ1}: Can SLMs generate empathetic responses in PTSD contexts comparable to large models?

\textbf{RQ2}: How does model size influence empathetic response quality, and is there a ``knowledge transfer ceiling'' for SLMs?

\textbf{RQ3}: Do demographic attributes (e.g., age, gender, trauma type) affect model performance in empathy generation?

\textbf{RQ4}: How well do standard automatic metrics (e.g., cosine similarity) align with human judgments of perceived empathy?

To pursue these questions, we address two key challenges: (1) the limited expressive capacity of SLMs, and (2) the scarcity of ethically sourced, trauma-informed dialogue data \citep{kokosi2022synthetic, bn2022privacy, bn23_interspeech}. We leverage TIDE (Trauma-Informed Dialogue for Empathy), a dataset of 10,000 two-turn conversations grounded in metadata derived from 500 diverse PTSD client personas, introduced in prior work focused on synthetic persona generation \citep{bn2025thousand}. While that work emphasized dataset construction, our study uses TIDE to evaluate whether small models can generate emotionally attuned, trauma-sensitive responses in therapeutic contexts. TIDE is designed to train models in real-time empathetic validation, recognizing emotion, normalizing distress, and offering supportive reflections within safe conversational bounds. We assess SLM performance using both automatic metrics and human evaluations of perceived empathy.

The remainder of this paper details our methodology, presents empirical results, and discusses the implications for building lightweight, emotionally aware AI systems for mental health support.

\vspace{-0.1cm}
\section{Motivation}
\vspace{-0.2cm}
PTSD affects over 13 million adults annually in the U.S. alone \cite{apa_ptsd_stats}, often impairing emotional regulation and increasing sensitivity to perceived invalidation. In such contexts, empathetic communication is critical, users seek acknowledgment, validation, and safety, not task completion. Yet, most conversational AI systems remain optimized for functionality: setting reminders, offering distractions, or providing information. When a user says ``I’m feeling overwhelmed,'' assistants may reply, ``Would you like to hear a podcast?,'' a utilitarian response that can feel dismissive or tone-deaf in emotionally vulnerable moments. While LLMs  show promise for empathetic generation, they are rarely trauma-specific and are too resource-intensive for scalable or privacy-preserving deployment. Small Language Models (SLMs) offer practical advantages, lower latency, on-device use, and reduced risk, but face three key gaps in the PTSD context:
\begin{itemize}
\vspace{-0.2cm}
\item Lack of fine-tuning on trauma-informed dialogue patterns, such as avoidance cues or dissociative disclosures.
\vspace{-0.2cm}
\item Absence of evaluation frameworks grounded in safety, empathy, and distress mitigation.
\vspace{-0.2cm}
\item Minimal attention to ethical design for simulated therapy or support contexts.
\vspace{-0.2cm}
\end{itemize}

This work addresses these gaps by creating the TIDE dataset, a clinically reviewed PTSD-specific synthetic corpus, and fine-tuning SLMs to respond empathetically. It combines automatic metrics with trauma-informed human evaluation to assess whether these SLM's can provide safe, meaningful support in sensitive mental health interactions.

\vspace{-0.1cm}
\section{Related Work}
\vspace{-0.2cm}
We survey six interconnected areas: fine-tuning and prompting (especially for small-scale models) \cite{1tahir2023cbt, 3raamkumar2024multidimensional, 4yang2024hef, 10hsu2023empathydialo, 12tiwari2023intentaware, 14ma2024emotionaldialogue, 35yu2024finetuning, li2024optimizing}, empathetic AI in mental health \cite{6sharma2022partner, 7chang2021support, 14ma2024emotionaldialogue, 16wu2023ptsd, sharma2020computational, zheng2021comae, wang2024patient, welivita2023empathetic}, emotion and intent modeling \cite{13su2021emotionintent, yang2024modeling}, reinforcement learning for empathy \cite{6sharma2022partner, 7chang2021support, 10hsu2023empathydialo}, human-centered evaluation \cite{sharma2020computational, xu2024multi, marrapese2024novel, park2024building}, and ethical safety in trauma-informed AI \cite{hua2024large, gabriel2024can, almakinah2024enhancing}. The following subsections review each strand in detail, while the summary highlights key gaps in trauma-sensitive, small-model fine-tuning and evaluation.

\vspace{-0.1cm}
\subsection{Fine-Tuning and Prompting for Empathetic Dialogue}
\vspace{-0.2cm}
Fine-tuning improves LLMs' empathetic response generation \cite{1tahir2023cbt, 3raamkumar2024multidimensional, 4yang2024hef, 10hsu2023empathydialo, 12tiwari2023intentaware, 14ma2024emotionaldialogue, 35yu2024finetuning, li2024optimizing}. \citet{1tahir2023cbt} fine-tuned small models (e.g., Mistral 7B, Qwen 2.5-7B, LLaMA -3.1-8B) for CBT interventions with ethical considerations, but without specific focus on PTSD. \citet{li2024optimizing} explored instruction tuning for general psychological counseling. Other work incorporates psychological frameworks (e.g., Chain of Empathy \cite{lee2023chain}), few-shot prompting \cite{qian2023harnessing, filienko2024toward}, or specialized tasks like cognitive distortion detection \cite{chen2023empowering}. Embedding fusion \cite{4yang2024hef, rasool2025emotion} and plug-and-play modules \cite{10hsu2023empathydialo} enhance general empathy on datasets like EmpatheticDialogues \cite{qian2023harnessing, loh2023harnessing}, but rarely address trauma-informed tuning or small-model safety controls. Our work directly fine-tunes SLMs on a PTSD-specific dataset (TIDE) and evaluates their limitations.

\subsection{Empathetic AI in Mental Health Contexts}
\vspace{-0.1cm}
Mental health AI has largely focused on general emotional support. Early work proposed computational empathy frameworks \cite{sharma2020computational} and empathetic response models \cite{zheng2021comae, welivita2023empathetic}. Other efforts include empathic rewriting for online forums \cite{6sharma2022partner}, supportive response selection \cite{7chang2021support}, and enhancing emotional richness \cite{14ma2024emotionaldialogue}. LLMs have been applied to PTSD diagnosis \cite{16wu2023ptsd} and patient simulation for clinical training \cite{wang2024patient}, but often use general datasets \cite{welivita2023empathetic} and lack PTSD-specific focus. Some chatbots have shown potential in mitigating social exclusion \cite{de2020effectiveness} and general support \cite{chin2023potential}, but comparisons reveal LLMs often underperform against human therapists \cite{almakinah2024enhancing} or trained peers in specialized contexts \cite{iftikhar2024therapy}.

\subsection{Emotion and Intent Recognition}
\vspace{-0.1cm}
Accurate modeling of user emotion and intent is essential for generating empathetic responses \cite{13su2021emotionintent, 12tiwari2023intentaware}. \citet{13su2021emotionintent} used multi-source perception, while \citet{12tiwari2023intentaware} integrated intent prediction into a T5 model, improving responses on general datasets. \citet{zheng2021comae} combined emotion and dialogue act classification in a hierarchical framework. \citet{yang2024modeling} applied Appraisal Theory to model alignment in online and therapy conversations. Though these methods enhance contextual understanding, few explicitly consider PTSD-specific emotions, triggers, or safeguards.
\vspace{-0.2cm}
\subsection{Reinforcement Learning for Empathy}
\vspace{-0.1cm}
RL has been applied to empathetic response generation. \citet{6sharma2022partner} trained a policy for empathic rewriting via human feedback, while \citet{7chang2021support} used RL to align responses with support strategies.  \citet{10hsu2023empathydialo} proposed modular empathy control. These methods are promising but rarely integrate clinical knowledge: PTSD symptoms or therapeutic boundaries, and RL for trauma-specific dialogue remains largely unexplored.
\vspace{-0.2cm}
\subsection{Human-Centered and Clinically-Grounded Evaluation}
\vspace{-0.1cm}
Recognizing the limits of automated metrics, recent work emphasizes human-centered evaluation. Approaches include A/B testing and subjective empathy ratings \cite{10hsu2023empathydialo, 12tiwari2023intentaware, 6sharma2022partner}, and more structured frameworks like EPITOME \cite{sharma2020computational, loh2023harnessing}, multidimensional empathy assessments \cite{xu2024multi, 3raamkumar2024multidimensional}, and psychotherapy-informed metrics (e.g., alignment, emotion consistency \cite{marrapese2024novel}). Safety-focused benchmarks are also emerging \cite{park2024building}. However, trauma-sensitive metrics for perceived safety, trust, or distress mitigation are still rare. Our study addresses this by using human ratings on PTSD-specific scenarios from the TIDE dataset.

\subsection{Ethical Considerations and Safety}
\vspace{-0.1cm}
Empathetic AI in mental health must consider safety, though many studies mention only general guardrails \cite{1tahir2023cbt, 4yang2024hef, 6sharma2022partner}. Some explore toxicity prevention or privacy techniques like federated learning \cite{almakinah2024enhancing}, but few address PTSD-specific risks (e.g., re-traumatization \cite{iftikhar2024therapy}, dissociation, crisis handling). Bias and equity issues have also emerged, with LLMs sometimes showing disparities in responses \cite{gabriel2024can}. Though concerns around dependency and misinterpreted empathy are acknowledged \cite{de2020effectiveness}, empirical solutions, like explicit AI disclaimers or boundary-aware tuning, are limited. Safety frameworks remain an open research area \cite{park2024building}.

\subsection{Summary}
\vspace{-0.1cm}
Significant progress has been made in empathetic dialogue using fine-tuning, prompting, RL, and intent modeling, with growing attention to evaluation and ethics. However, key gaps remain:
\begin{itemize}
\vspace{-0.2cm}
\item Lack of trauma-specific fine-tuning for SLMs incorporating PTSD symptom patterns and avoidance strategies.
\vspace{-0.2cm}
\item Absence of standardized, trauma-informed evaluation for safety and distress mitigation.
\vspace{-0.2cm}
\item Limited ethical design tailored to trauma therapy simulation or support.
\vspace{-0.2cm}
\end{itemize}
Our work aims to address these gaps by fine-tuning small language models using the novel, PTSD-specific TIDE dataset, evaluating their performance with human-centered, trauma-informed metrics derived from realistic scenarios, and grounding our approach in the ethical considerations necessary for these sensitive interactions.

\section{Methodology}
\vspace{-0.2cm}

\begin{figure*}
    \centering
    \includegraphics[width=\linewidth]{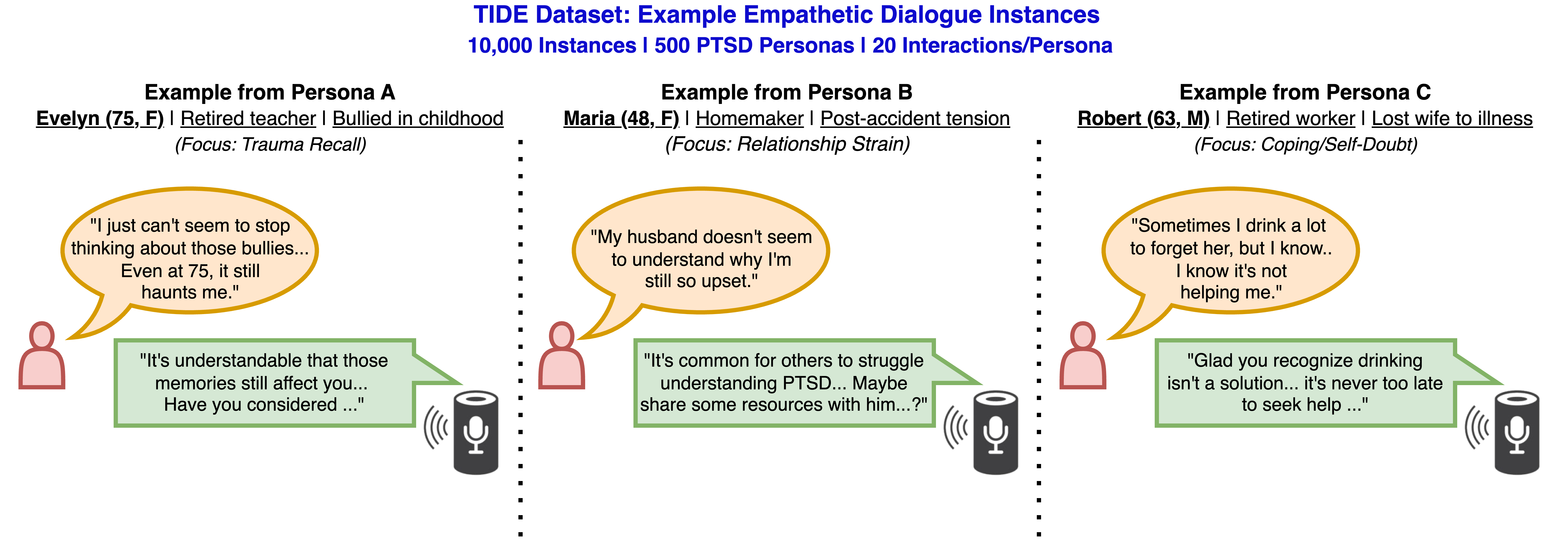}
    \caption{Illustrative examples of the two-turn dialogue structure within the TIDE dataset, showcasing synthetic empathetic interactions across diverse PTSD personas. Each instance consists of a \texttt{user\_input} reflecting a trauma-related challenge and a corresponding \texttt{reference\_response} providing empathetic validation. The examples represent distinct trauma themes (e.g., trauma recall, relationship strain, coping/self-doubt), with persona metadata highlighting the breadth of contextual grounding used to guide generation. TIDE comprises 10{,}000 such instances derived from 500 unique personas (20 per persona), with reference responses generated using Claude Sonnet 3.5. Example dialogues reviewed and validated by PTSD clinical specialists.}
    \label{fig:tide_dataset}
\end{figure*}

\subsection{Data Description: TIDE Dataset}
\vspace{-0.1cm}
Our empirical evaluations use the Trauma-Informed Dialogue for Empathy (TIDE) dataset, a novel large-scale resource designed to assess and improve the empathetic capabilities of small language models in trauma-informed contexts.

\subsection{Dataset Composition and Structure}
\vspace{-0.1cm}
TIDE consists of 10,000 two-turn dialogues derived from 500 diverse client personas, each grounded in characteristics associated with PTSD \citep{bn2025thousand}. For each persona, 20 distinct interaction scenarios were created to reflect varying facets or moments in their trauma narrative, while remaining thematically coherent. All scenarios were clinically reviewed by co-author psychologists specializing in PTSD and Prolonged Exposure (PE) therapy to ensure emotional plausibility and trauma sensitivity. Each instance in TIDE includes:

\textbf{Client Utterance (\texttt{user\_input})}: A realistic expression of distress, PTSD symptoms (e.g., avoidance, intrusive memories, self-blame, relational struggles), or emotionally complex states tied to the persona (e.g., Persona 1).

\textbf{Empathetic Response (\texttt{reference\_response})}: A gold-standard empathetic reply designed to validate and support, not diagnose or treat. These responses emphasize emotional connection (e.g., ``It's understandable that those memories still affect you. Your feelings are valid. Have you considered talking to a therapist who specializes in trauma?''). See Fig.~\ref{fig:tide_dataset} for sample dialogues.

\textbf{Generation Methodology:} The reference responses (and potentially client utterances; pending internal confirmation) were generated using Claude Sonnet 3.5 \citep{anthropic2024claude}. Prompts included persona background and user utterance to elicit contextually appropriate, highly empathetic replies aligned with supportive communication principles.

\textbf{Application \& Availability:} In our evaluations, SLMs are prompted with \texttt{user\_input} from TIDE, and their outputs (zero-shot or fine-tuned) are assessed against the corresponding \texttt{reference\_response} using multiple metrics. The full dataset, user inputs, reference responses, and persona identifiers, is planned for public release to support broader empathetic AI research.

\vspace{-0.15cm}
\subsection{Metrics and Analysis}
\label{sec:metrics_analysis}
\vspace{-0.1cm}
To evaluate the empathetic response generation capabilities of the Small Language Models (SLMs) under investigation, both prior to (zero-shot) and subsequent to fine-tuning on the TIDE dataset, we employ a multi-faceted evaluation strategy. This strategy integrates established automatic metrics for quantifiable assessment of text quality and similarity, alongside rigorous human evaluations designed to capture the critical, yet often elusive, perceived quality of empathy.

\vspace{-0.2cm}
\subsubsection{Automatic Evaluation Metrics}
\label{sec:auto_metrics}
\vspace{-0.1cm}

To quantitatively assess the quality of generated responses, we use both reference-based and reference-free automatic metrics, including BERTScore \cite{zhang2019bertscore}, ROUGE-L \cite{lin2004rouge}, METEOR \cite{banerjee2005meteor}, Distinct-1/2 \cite{li2016diversity}, and cosine similarity using \textit{all-MiniLM-L6-v2} embeddings \cite{allMiniLML6v2}. These metrics collectively capture semantic similarity, lexical overlap, fluency, and response diversity, and are widely used for evaluating chatbot and summarization outputs. We report results across all models in both zero-shot and fine-tuned configurations (see Table~\ref{tab:combined_results_multiple_metrics}).

While these metrics offer scalable benchmarks for response quality, they have significant limitations in evaluating empathy and clinical appropriateness. Automatic metrics primarily capture surface-level textual similarity and fluency but may not adequately reflect the subtle aspects of perceived empathy, such as emotional resonance, validation, and contextual appropriateness in tone. To address this, we conducted a human evaluation (N=116) to directly assess empathetic quality (see Appendix~\ref{sec:QualtricsSurvey} for the survey, Appendix~\ref{sec:human_evaluation_protocol} for the protocol, and Appendix~\ref{sec:human_study_recruitment} for recruitment).

\vspace{-0.15cm}
\section{Human Evaluation Results}
\label{sec:human_eval_results}
\vspace{-0.2cm}
Given that the TIDE examples were reviewed for clinical realism by PTSD experts, we designed the human evaluation to focus specifically on perceived empathy, trusting that the input scenarios reflected realistic trauma narratives.
We conducted a controlled human evaluation to compare the empathetic quality of model responses across two conditions: zero-shot and fine-tuned states. Participants rated perceived empathy on a 5-point Likert scale for each generated response, following the protocol outlined in Appendix~\ref{sec:human_evaluation_protocol}. To systematically assess the impact of fine-tuning, we performed a pairwise comparison for each model and input scenario. For each of the six conversational scenarios and eight models, we computed the mean empathy score across participants and assigned a win to the condition (ZS or FT) with the higher average rating. Additionally, we benchmarked each model's output against the ground-truth reference from Claude Sonnet 3.5, awarding two points if Claude outperformed both ZS and FT outputs, one point if it outperformed only one, and zero otherwise.

Table~\ref{tab:human_eval_summary} summarizes model wins across scenarios, highlighting the benefits of fine-tuning and the idealized benchmark performance established by Claude Sonnet 3.5. We observe that: (1) ZS models beat FT in 26 cases, (2) FT beat ZS in 22, and (3) Sonnet 3.5 dominated 72/96 comparisons.

Across six scenarios, ZS models achieved wins in 2  scenarios, FT models in two scenarios, with two scenarios resulting in a tie.

These results indicate an equal number of scenario wins across ZS and FT. However, fine-tuning gains were not universal; ZS sometimes won scenarios by a higher margin (e.g., Scenario 4). 

Claude Sonnet 3.5 consistently outperformed all small models across most scenarios, reaffirming the quality gap between frontier models and resource-efficient architectures. Nevertheless, the relatively close margins in several cases (e.g., FT Llama 3.2-1B and DeepSeek-R1-Distill-Qwen 1.5B) suggest that small models can approach high-quality empathetic responses with targeted adaptation.

Overall, human evaluation results corroborate our automatic metric findings while also revealing that empathetic improvement through fine-tuning is scenario-dependent and non-monotonic, especially for smaller parameter models.

\subsection{Human Evaluation Visualization}
\label{sec:human_eval_visualization}
\vspace{-0.1cm}

\begin{figure*}[t] \centering \includegraphics[width=\linewidth]{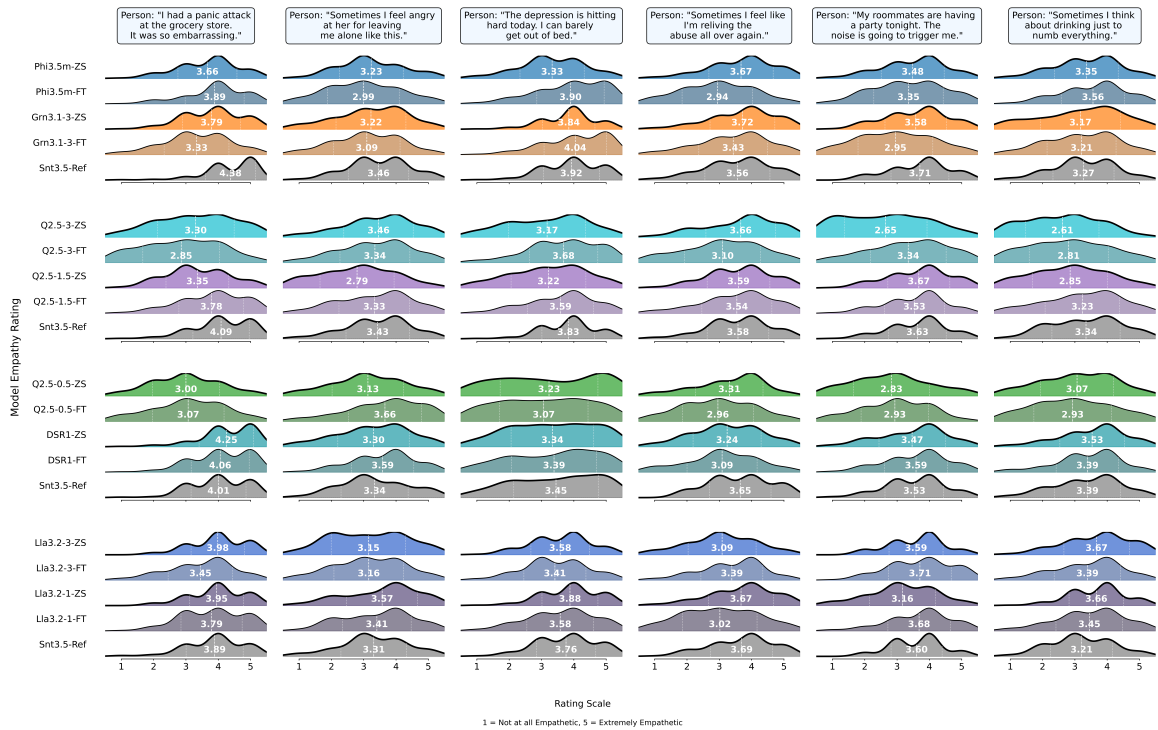} \caption{Distribution of human empathy ratings (1 = Not at all empathetic, 5 = Extremely empathetic) for zero-shot (ZS), fine-tuned (FT), and reference (Claude Sonnet 3.5) responses across six conversational scenarios. Means are annotated on each distribution. Fine-tuning generally shifts model responses toward higher perceived empathy, although improvements vary by model and scenario.} \label{fig:empathy_ridgeline} \end{figure*}

To complement the aggregate win-loss analysis, we visualize the full distribution of empathy ratings for each model across all conversational scenarios in Figure~\ref{fig:empathy_ridgeline}. The figure presents kernel density plots of participant Likert ratings (1-5 scale), grouped by model (ZS, FT, and reference) and input scenario.

\vspace{-0.15cm}
\paragraph{Key Trends in Empathy Ratings:}
\begin{itemize}
\vspace{-0.2cm}
    \item \textbf{Fine-tuning gains are scenario- and size-dependent:}
    Larger models (e.g., \textit{Phi 3.5B}, \textit{Llama 3.2-3B}, \textit{Qwen 2.5-3B}) improve most in complex scenarios (3, 5, 6). Smaller models (e.g., \textit{Qwen 2.5-0.5B}, \textit{Llama 3.2-1B}) show gains mainly in simpler ones (1, 2, 5), while mid-sized models (e.g., \textit{Qwen 2.5-1.5B}, \textit{R1-Distill-1.5B}) perform well in both. Scenarios 3 and 5 are most responsive overall.
    \vspace{-0.2cm}
    \item \textbf{Claude Sonnet 3.5 remains the top performer:}
    It consistently receives the highest empathy ratings, though in scenarios like 3 and 4, fine-tuned SLMs narrow the gap.
    \vspace{-0.2cm}
    \item \textbf{Participant variability is high:}
    Empathy ratings vary widely across individuals, underscoring the subjectivity of perceived empathy and the value of aggregate comparisons.
    \vspace{-0.2cm}
\end{itemize}

Figure~\ref{fig:empathy_ridgeline} visually reinforces the finding that the improvement due to fine-tuning varies based on both model architecture and conversational context.

\begin{figure*}
    \centering
    \includegraphics[width=0.93\linewidth]{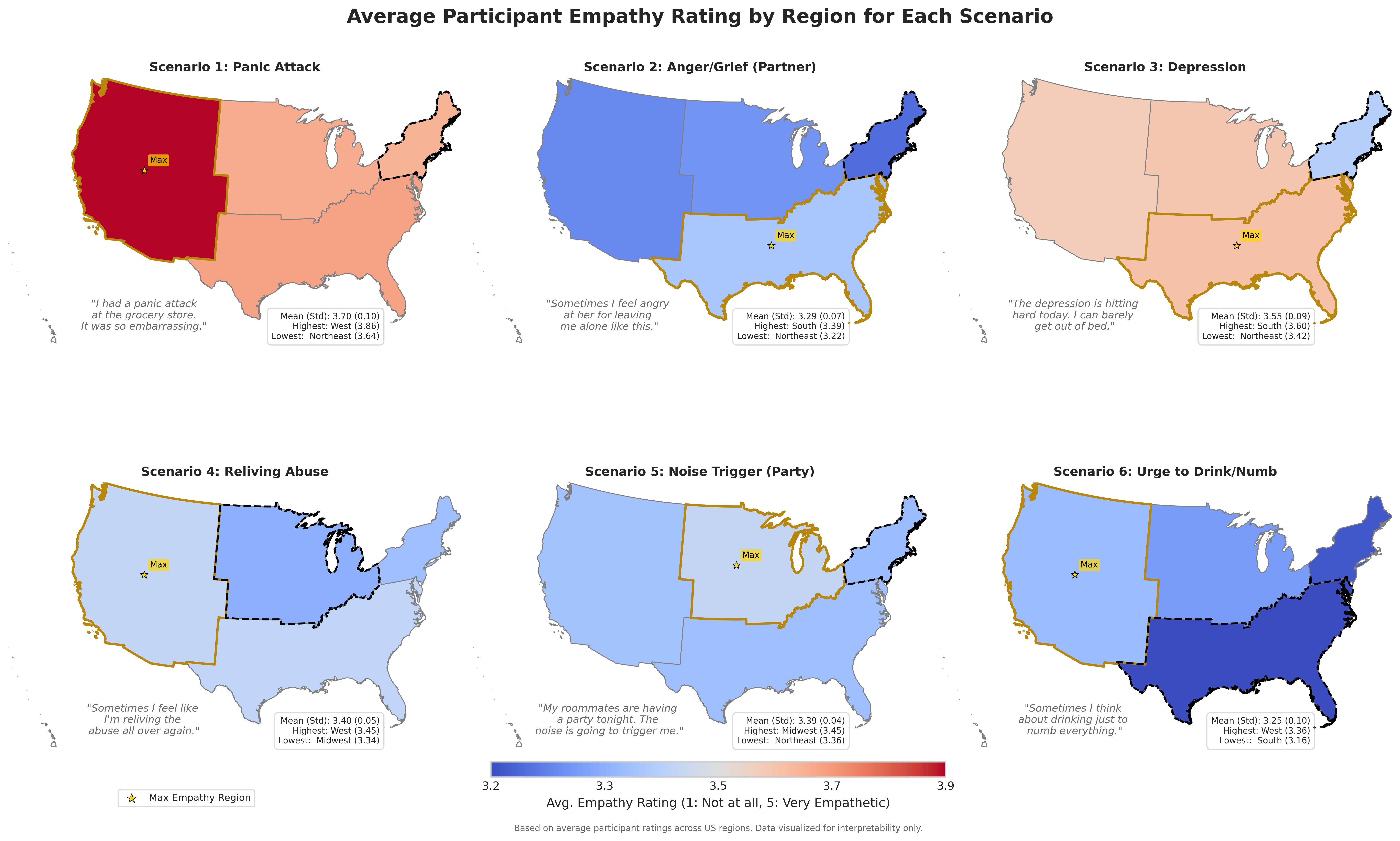}
    \caption{Average participant empathy ratings by U.S. region across six mental health scenarios.
Each map depicts mean empathy ratings (1 = Not at all, 5 = Very empathetic) reported by participants in response to a scenario-specific chatbot message. Color gradients indicate regional averages, with darker shades reflecting higher empathy. The region with the highest rating for each scenario is marked with a $\star$. Quotes illustrate representative messages used in the study. Ratings were averaged by U.S. Census regions for interpretability. West regions consistently reported higher empathy for acute distress (e.g., panic attack), while the South and Midwest showed stronger responses in other contexts (e.g., depression, grief).}
    \label{fig:map}
\end{figure*}

\begin{table*}[!htbp]
    \centering
    \scriptsize
    \caption{Comparison of Zero-Shot (ZS) and Fine-Tuned (FT) Performance of SLMs on the TIDE Dataset across multiple metrics. Values are Mean $\pm$ Std. Dev., rounded to two and one decimal places respectively. FT models were trained on 6,000 samples and evaluated on 4,000 samples over 3 epochs (batch size = 2, gradient accumulation = 4, learning rate = 1e-5, LoRA rank = 8, LoRA $\alpha$ = 16, max length = 80).}
    \label{tab:combined_results_multiple_metrics}
    \vspace{-0.2cm}
    \resizebox{\textwidth}{!}{
    \begin{tabular}{l
                    cc cc cc cc cc cc}
        \toprule
        \textbf{Model Name} & 
        \multicolumn{2}{c}{\textbf{Cosine Sim.}} & 
        \multicolumn{2}{c}{\textbf{BERTScore F1}} &
        \multicolumn{2}{c}{\textbf{METEOR}} &
        \multicolumn{2}{c}{\textbf{ROUGE-L}} &
        \multicolumn{2}{c}{\textbf{Distinct-1}} &
        \multicolumn{2}{c}{\textbf{Distinct-2}} \\
        \cmidrule(lr){2-3} \cmidrule(lr){4-5} \cmidrule(lr){6-7} \cmidrule(lr){8-9} \cmidrule(lr){10-11} \cmidrule(lr){12-13}
         & \textbf{ZS} & \textbf{FT} & \textbf{ZS} & \textbf{FT} & \textbf{ZS} & \textbf{FT} & \textbf{ZS} & \textbf{FT} & \textbf{ZS} & \textbf{FT} & \textbf{ZS} & \textbf{FT} \\
        \midrule
        Phi 3.5-mini-instruct       & 0.56 $\pm$ 0.2 & 0.58 $\pm$ 0.2 & 0.88 $\pm$ 0.0 & 0.89 $\pm$ 0.0 & 0.32 $\pm$ 0.1 & 0.33 $\pm$ 0.1 & 0.83 $\pm$ 0.0 & 0.84 $\pm$ 0.0 & 0.19 $\pm$ 0.0 & 0.21 $\pm$ 0.0 & 0.33 $\pm$ 0.0 & 0.35 $\pm$ 0.0\\
        granite 3.1-3b             & 0.57 $\pm$ 0.2 & 0.58 $\pm$ 0.2 & 0.88 $\pm$ 0.0 & 0.89 $\pm$ 0.0 & 0.33 $\pm$ 0.1 & 0.34 $\pm$ 0.1 & 0.82 $\pm$ 0.1 & 0.83 $\pm$ 0.1 & 0.18 $\pm$ 0.0 & 0.20 $\pm$ 0.0 & 0.32 $\pm$ 0.0 & 0.35 $\pm$ 0.0\\
        Qwen 2.5-3B                 & 0.55 $\pm$ 0.2 & 0.56 $\pm$ 0.2 & 0.89 $\pm$ 0.0 & 0.90 $\pm$ 0.0 & 0.31 $\pm$ 0.1 & 0.33 $\pm$ 0.1 & 0.83 $\pm$ 0.1 & 0.84 $\pm$ 0.1 & 0.20 $\pm$ 0.0 & 0.24 $\pm$ 0.0 & 0.35 $\pm$ 0.0 & 0.37 $\pm$ 0.0\\
        Qwen 2.5-1.5B               & 0.57 $\pm$ 0.2 & 0.59 $\pm$ 0.2 & 0.89 $\pm$ 0.0 & 0.90 $\pm$ 0.0 & 0.33 $\pm$ 0.1 & 0.34 $\pm$ 0.1 & 0.83 $\pm$ 0.1 & 0.84 $\pm$ 0.1 & 0.21 $\pm$ 0.0 & 0.25 $\pm$ 0.0 & 0.35 $\pm$ 0.0 & 0.37 $\pm$ 0.0\\
        Qwen 2.5-0.5B               & 0.50 $\pm$ 0.2 & 0.52 $\pm$ 0.2 & 0.88 $\pm$ 0.0 & 0.88 $\pm$ 0.0 & 0.26 $\pm$ 0.1 & 0.27 $\pm$ 0.1 & 0.80 $\pm$ 0.1 & 0.81 $\pm$ 0.1 & 0.17 $\pm$ 0.0 & 0.21 $\pm$ 0.0 & 0.30 $\pm$ 0.0 & 0.31 $\pm$ 0.0\\
        R1-Distill-Qwen 1.5B      & 0.60 $\pm$ 0.2 & 0.61 $\pm$ 0.2 & 0.89 $\pm$ 0.0 & 0.90 $\pm$ 0.0 & 0.35 $\pm$ 0.1 & 0.36 $\pm$ 0.1 & 0.80 $\pm$ 0.1 & 0.81 $\pm$ 0.1 & 0.22 $\pm$ 0.0 & 0.23 $\pm$ 0.0 & 0.35 $\pm$ 0.0 & 0.39 $\pm$ 0.0\\
        Llama 3.2-3B              & 0.61 $\pm$ 0.2 & 0.63 $\pm$ 0.2 & 0.89 $\pm$ 0.0 & 0.90 $\pm$ 0.0 & 0.35 $\pm$ 0.1 & 0.36 $\pm$ 0.1 & 0.79 $\pm$ 0.0 & 0.79 $\pm$ 0.0 & 0.20 $\pm$ 0.0 & 0.25 $\pm$ 0.0 & 0.33 $\pm$ 0.0 & 0.37 $\pm$ 0.0\\
        Llama 3.2-1B              & 0.62 $\pm$ 0.2 & 0.64 $\pm$ 0.2 & 0.89 $\pm$ 0.0 & 0.90 $\pm$ 0.0 & 0.37 $\pm$ 0.1 & 0.38 $\pm$ 0.1 & 0.80 $\pm$ 0.0 & 0.80 $\pm$ 0.0 & 0.20 $\pm$ 0.0 & 0.22 $\pm$ 0.0 & 0.32 $\pm$ 0.0 & 0.35 $\pm$ 0.0\\
        \bottomrule
    \end{tabular}
    }
\end{table*}

\vspace{-0.1cm}
\section{Discussion}
\label{sec:discussion}
\vspace{-0.15cm}
Our findings reveal interesting insights into the empathetic capabilities of small language models (SLMs) and the effects of targeted fine-tuning on trauma-informed dialogue generation.
\subsection{Fine-Tuning Improves Empathy, Unevenly}
\vspace{-0.15cm}
Across automatic metrics (cosine similarity, BERTScore, ROUGE-L, Meteor, Distinct-1/2) and human evaluation, fine-tuning improved metric performance over the ZS baseline but not uniformly. As detailed in Section~\ref{sec:human_eval_results}, FT models outperformed their ZS counterparts in 22 cases, while ZS models outperformed FT models in 26 cases. Smaller models (e.g., Qwen 2.5-0.5B, Llama 3.2-1B) showed greater variability, sometimes producing more natural or emotionally resonant responses without fine-tuning. This suggests that fine-tuning these small models may overfit them to affirmation-heavy styles in synthetic data, dampening nuanced emotional calibration in some contexts.

\vspace{-0.2cm}
\subsection{Knowledge Transfer Ceiling in Empathy}
\vspace{-0.2cm}
Even after fine-tuning, a persistent performance gap remained between small models and the Claude Sonnet 3.5 reference. Claude consistently achieved the highest human empathy ratings, dominating 72 out of the 96 direct comparisons (75\%). The Qwen models including the 0.5B one often approached and beat Claude's emotional resonance, especially for complex or layered emotional inputs (e.g., Scenario 2, involving anger and isolation). However, this was not the case for the other larger models from Phi, Granite, and Llama. These findings suggest a ``knowledge transfer ceiling'': although fine-tuning boosts empathetic response quality, models with 0.5-3B parameters show clear limits in replicating the contextually rich, emotionally sensitive language of the frontier model.

\vspace{-0.2cm}
\subsection{Automatic Metrics: Incomplete Proxies}
\vspace{-0.1cm}
While FT models improved on metrics such as cosine similarity and BERTScore, these gains did not always match higher human-perceived empathy. Models with similar similarity scores often differed in participant ratings, underscoring the limits of purely similarity-based evaluation for affective tasks. Metrics like ROUGE and Distinct aligned even less with human judgments, reflecting their focus on surface overlap rather than emotional quality. This underscores the need for human evaluation in affective generation tasks and for developing more emotionally aware automatic metrics.

\vspace{-0.1cm}
\subsection{Scenario-Specific Sensitivity}
\vspace{-0.1cm}
Analysis across six scenarios revealed that model improvements were highly input-sensitive. Scenarios involving clear expressions of distress (e.g., ``I had a panic attack at the grocery store'') showed greater fine-tuning gains, while more complex emotional situations (e.g., ``I feel angry at her for leaving me'') elicited less consistent improvements. This suggests that fine-tuning with trauma-informed data helps models handle direct expressions of suffering but leaves room for improvement in handling nuanced, multi-layered emotions like anger, guilt, or relational tension. Future fine-tuning strategies might benefit from more diverse emotional curricula that explicitly balance affirmation with context-sensitive complexity.

\vspace{-0.2cm}
\subsection{User Diversity in Empathy Preferences}
\label{sec:contextual_and_user}
\vspace{-0.1cm}
Empathy is not one-size-fits-all. Our demographic analysis revealed significant variation in how different user groups perceived empathy in model responses, highlighting the importance of considering diversity when developing and evaluating systems trained on synthetic data. Yet, certain response characteristics, emotional clarity, contextual relevance, and actionable support, consistently emerged as valued across groups.
\vspace{-0.05cm}

\vspace{-0.2cm}
\paragraph{Older Adults: Reassurance Before Resolution}
Participants over 60 consistently rated responses from select small models higher than younger participants. Notably, Llama 3.2-3B showed a statistically significant uplift among older adults (\textit{mean = 3.66}, \textit{SD = 0.94}) compared to younger ones (\textit{mean = 3.51}, \textit{SD = 1.01}; \textit{t} = 2.89, \textit{p} = 0.004, \textit{Cohen’s d} = 0.16). Similarly, Phi 3.5 Mini was also rated more empathetic by older adults (\textit{mean = 3.66}) than younger adults (\textit{mean = 3.48}; \textit{p} = 0.002), with a moderate effect size. These models excelled not because of their size, but due to their emotionally steady delivery. For example, Llama 3.2-1B ZS offered, \textit{``It’s nothing to be embarrassed about... Would you like to try a grounding technique?''}, and Qwen 2.5-1.5B FT reassured users with, \textit{``Absolutely... It's not your fault... Try sitting by a window… You can do this.''} These responses validated emotion, normalized experience, and offered simple coping strategies. In contrast, lower-rated models like Qwen 2.5 0.5B FT and ZS produced vague or awkward phrasing (e.g., \textit{``Human experience can be frustrating''}, or the unclear phrase \textit{``Humanning the past''}) or introduced irrelevant content. This indicates that older users favor responses that are emotionally anchored and practical, suggesting design priorities should lean toward clarity and support over abstraction.

\vspace{-0.2cm}
\paragraph{Education: Empathy + Practicality?}
\vspace{-0.1cm}
Graduate-educated participants favored responses balancing emotional validation with actionable suggestions. For example, Claude Sonnet 3.5 received a high average empathy rating of 4.48 from graduate users in a high-stress scenario, though the difference was not statistically significant (\textit{p} = 0.28). Robust differences emerged for some models: Qwen 2.5-3B and Qwen 2.5-1.5B were rated significantly lower by graduate participants than non-graduates (\textit{p} < 0.001; Cohen’s $d$ = -0.26 and -0.20, respectively), suggesting sensitivity to vague or generic phrasing. Conversely, while Phi 3.5 Mini showed no overall difference across education groups (\textit{p} = 0.96), one response for Prompt Q17 trended toward higher ratings among graduates (\textit{p} = 0.088, Cohen’s $d$ = 0.26), indicating a preference for steady, well-calibrated delivery in emotionally complex situations. These findings suggest educational background can shape empathy perception in nuanced edge cases, underscoring the need for adaptable, audience-aware generation.

\vspace{-0.15cm}
\paragraph{Gender: Tone Matters, But So Does Substance.}
Despite common assumptions of divergent gender-based preferences, empathy ratings from men and women showed strong alignment. Across all evaluated models, including top performers like Claude Sonnet 3.5 and Llama 3.2-3B (ZS), no statistically significant differences emerged across gender groups (\textit{p} > 0.15 for all), and effect sizes were negligible ($|d|$ < 0.05). While qualitative impressions hinted at tonal preferences, such as women slightly favoring gentler framings (e.g., DeepSeek-R1 ZS) and men leaning toward directive tones (e.g., DeepSeek-R1 FT), these patterns were not reflected in aggregate empathy scores. Crucially, the Qwen 2.5 variants were \textit{rated consistently lower} by both men and women, with no significant differences in mean ratings across gender (\textit{p} = 0.37-0.99; Cohen’s $d$ < 0.05). For example, Qwen 2.5-3B received average ratings of 3.14 (men) and 3.20 (women), underscoring shared expectations for emotional clarity and coherence in therapeutic contexts, regardless of gender identity.

\paragraph{Empathy Map: It looks different from here!}
Figure~\ref{fig:map} shows region-specific variation in empathy ratings across mental health scenarios. For acute distress (e.g., panic, abuse), participants in the West gave significantly higher ratings than other regions (Mean = 3.75; F(3, N) = 3.51, p = 0.015), suggesting heightened sensitivity to emotional urgency. In contrast, for depression and grief, the South (3.62) and Midwest (3.46) rated responses more empathetically, though differences were not significant (F = 0.78, p = .505). While our study does not examine causal mechanisms, these findings point to the need for culturally and regionally sensitive design in empathetic AI.

\vspace{-0.15cm}
\paragraph{Takeaway:}
Although principles such as empathetic communication, emotional validation, contextual grounding, and clear next steps are valued, their delivery must be sensitively adapted to user profiles. Designing for empathy requires systems to be both context- and user-aware, accounting for scenario content \& individual differences in how comfort, care, and counsel are received.

\vspace{-0.15cm}
\subsection{One Model to Rule Them All?}
\vspace{-0.1cm}
Our analysis suggests that no single model excels across all emotional scenarios. Empathy perception is highly context-dependent, and different models perform best in different situations. In Scenario 1 (panic attack), Claude Ref-QType1 led with a mean of 4.38, but DeepSeek-R1 ZS (4.25) and FT (4.06) were close behind, showing that both reference and instruction-tuned models resonated well for acute anxiety. In Scenario 2 (grief/anger), Llama 3.2-1B ZS (3.57) and Qwen 2.5 0.5B FT (3.65) outperformed larger models, possibly due to their concise, emotionally direct responses, highlighting that smaller models can still deliver contextually attuned empathy. Scenario 3 (depression) favored Granite -3.1 FT (4.04), Claude Ref-QType1 (3.92), and Llama 3.2-1B ZS (3.88), with top responses often emphasizing small, achievable actions, well-matched to the scenario’s emotional tone. In Scenario 4 (reliving abuse), Granite -3.1 ZS (3.72), Llama 3.2-1B ZS (3.67), and Claude Ref-QType4 (3.69) stood out. Grounding-focused, emotionally cautious responses likely drove effectiveness. In Scenario 5 (social anxiety), Llama 3.2-3B FT (3.71), Llama 1B FT (3.68), and Claude Ref-QType1 (3.71) led, suggesting fine-tuned Llama models were especially effective at offering relatable coping strategies. In Scenario 6 (substance use), Llama 3.2-3B ZS (3.67), 1B ZS (3.66), and DeepSeek-R1 ZS (3.53) performed best, balancing emotional validation with gentle redirection. These patterns indicate empathetic quality depends on more than model size or training, and contextual appropriateness matters most, which raises the question: should we build one model for all scenarios or multiple models fine-tuned to different emotional needs?

\vspace{-0.1cm}
\subsection{Do Perceptions of Empathy Vary by Race?}
\vspace{-0.1cm}
As shown in Table~\ref{tab:demographics}, 65.5\% of participants identified as White or Caucasian, and 19.8\% as Black or African American. Due to limited representation across other racial identities, we adopted a simplified grouping of White vs. non-White participants for exploratory analysis. While this binary comparison flattens cultural distinctions \citep{sap-etal-2022-annotators, may-etal-2019-measuring}, it allowed us to detect broad trends while maintaining statistical viability.

Across models, participants identifying as non-White consistently assigned higher empathy ratings. Claude Sonnet 3.5 received a mean of 3.81 from non-White participants versus 3.53 from White participants; similar trends appeared for Granite -3.1 ZS (3.68 vs. 3.49) and DeepSeek-R1 ZS (3.56 vs. 3.50). Even lower-rated models, such as the Qwen variants, showed this uplift. While not statistically conclusive, the consistency suggests that empathy perception is shaped not just by model output, but also by the recipient’s lived experience. To reduce linguistic ambiguity, all participants were native English speakers (Appendix~\ref{sec:human_evaluation_protocol}). While this control strengthened internal validity, it also centers the study in a Western, English-speaking context. Future work should examine these trends with more racially and linguistically diverse samples, and explore culturally empathetic response strategies.

\vspace{-0.15cm}
\subsection{Clinical Validation by Experts}
\vspace{-0.1cm}
The TIDE dataset underwent rigorous multi-stage validation led by a licensed clinical psychologist specializing in PTSD and PE therapy. This collaboration shaped the dataset from inception, not post-hoc. The process included:
\begin{enumerate}
\vspace{-0.2cm}
    \item \textbf{Persona Co-Design:} 500 diverse client personas were co-designed with the clinical expert to capture realistic trauma narratives and varied PTSD presentations.
    \vspace{-0.2cm}
    \item \textbf{Prompt Engineering and Refinement:} Dialogue-generation prompts were iteratively co-written and refined to embed trauma-informed practices, prioritizing emotional validation before offering suggestions.
    \vspace{-0.2cm}
    \item \textbf{Iterative Auditing:} Generated dialogues were repeatedly audited to adjust for safety, emotional appropriateness, and tonal realism.
    \vspace{-0.2cm}
\end{enumerate}

\vspace{-0.75cm}
\section{Future Work}
\vspace{-0.2cm}
We plan to use TIDE to train and evaluate models for therapeutic goals such as supporting junior therapists and enabling broader clinical applications. A key next step is to test whether differences between real and synthetic conversations (e.g.,~\citet{bn2025real}) correlate with patient outcomes like distress reduction or recovery, or at least confirm they do not harm therapeutic effectiveness, thereby clarifying the practical value of synthetic data in clinical settings. Our findings also reveal a trade-off between empathy and informativeness; future work should develop methods to balance immediate empathetic validation with deeper, more informative responses. Our FT approach can be compared with knowledge distillation to explore ways of transferring the empathetic capabilities of LLM's (ex. Claude~3.5) to smaller, more efficient architectures.

\vspace{-0.15cm}
\section{Conclusions}
\vspace{-0.2cm} 
This study demonstrates that small language models (0.5B-5B parameters) can be substantially improved to generate empathetic responses for PTSD support contexts through targeted fine-tuning on trauma-informed synthetic data. Using the TIDE dataset and a multi-pronged evaluation strategy, we show that fine-tuning boosts automatic similarity metrics and enhances human-perceived empathy, although gains are not uniform across models or scenarios. Our findings emphasize that small, resource-efficient models can meaningfully engage in emotionally supportive dialogue when carefully adapted, offering promising pathways for scalable, privacy-preserving mental health support systems.

\vspace{-0.2cm}
\section{Acknowledgement}
\vspace{-0.1cm}
This work was supported by the National Science Foundation (NSF) under Grant No. 2326144. Any opinions, findings, and conclusions or recommendations expressed are those of the author(s) and do not necessarily reflect the views of the NSF.

\vspace{-0.2cm}
\section{Data Availability}
\vspace{-0.15cm}
The TIDE dataset (Trauma-Informed Dialogue for Empathy) is publicly available at \url{https://huggingface.co/datasets/yenopoya/TIDE} under the CC BY-NC 4.0 license.
\vspace{-0.2cm}
\section{Limitations}
\vspace{-0.2cm}
Although small fine-tuned models can support empathetic generation, several limitations remain: (1) the synthetic TIDE dataset, while clinically validated, is derived from a single frontier model (Claude Sonnet 3.5) and may not capture the full complexity of real-world interactions; future work should integrate clinical transcripts and cross-cultural data; (2) our evaluation is confined to two-turn dialogues, which is the atomic unit of validation, but does not capture the longitudinal nature of therapeutic conversations; (3) human evaluation used a single 5-point Likert scale for perceived empathy and Prolific raters who may not represent clinical populations, limiting generalizability; (4) smaller models produced generic or overly affirmative replies, with ``Have you considered joining a support group'' appearing over ten times across conditions, indicating surface-level rather than context-specific empathy; and (5) while TIDE was clinically reviewed for safety, we did not conduct adversarial testing to probe harmful advice or mishandling of crisis situations, and we evaluated perceived empathy but not behavioral outcomes such as distress reduction or trust, which future work should assess.

\bibliography{custom}

\appendix
\onecolumn
\section{Survey Scenarios for Empathy Rating Study}
\label{sec:QualtricsSurvey}
\noindent
\textbf{Note:} Each set of four questions (e.g., Q.1-Q.4) corresponds to a single scenario as shown in Figure~\ref{fig:empathy_ridgeline}. The model-to-response mappings are provided in the main text and in the figure. The options were randomized in the survey. Please refer to Table \ref{tab:model-mapping} for the question/option mapping.

\small
\begin{quote}
\begin{Verbatim}[breaklines=true, breakanywhere=true]

Q.1:
Person: "I had a panic attack at the grocery store. It was so embarrassing."
Question 1 (Row 2619, Pair 1):
Please rate each response on empathy on a scale from 1 to 5, where 1 means 'Not at all empathetic' and 5 means 'Extremely empathetic'.:
Response 1: "Panic attacks can be really scary and overwhelming. Remember, it's not your fault. Have you informed your workplace about your PTSD? They might be able to offer some support."
Response 2: "Panic attacks can be really scary and overwhelming. Please don't be hard on yourself. Have you informed your workplace about your condition? They might be able to provide some support."
Response 3: "Panic attacks can be scary and overwhelming, but you're not alone. Have you learned any coping strategies for public emergencies? That could help in such situations in the future."
Response 4: "Panic attacks can be embarrassing, but remember, they're not your fault. Have you considered explaining to your employer about your PTSD? They might be understanding."
Response 5: "I'm sorry that happened. Panic attacks can be really scary and overwhelming. Please know it's not your fault, and many people experience them. Have you learned any grounding techniques that might help in those moments?"


Q.2:
Person: "I had a panic attack at the grocery store. It was so embarrassing."
Please rate each response on empathy on a scale from 1 to 5, where 1 means 'Not at all empathetic' and 5 means 'Extremely empathetic'.:
Response 1: "average person would have done that. You're strong for knowing when to seek help. Have you learned any grounding techniques to use when you feel the urge to use? They can be really helpful. I'd be happy to provide some resources if you're interested."
Response 2: "introverted guy like you, it can be isolating. But remember, panic attacks are a common symptom of PTSD. Have you considered joining a support group for people with similar experiences? It might help you feel less alone."
Response 3: "adult-onset PTSD can be really scary, but it'sNot at all embarrassing. Panic attacks are a common symptom of PTSD, especially for those who've experienced physical abuse. You did your best in a difficult situation. Have you learned any grounding techniques to help manage these episodes? They can be really"
Response 4: "absolutely, panic attacks can be scary and overwhelming. It's not your fault. Next time, try sitting by a window or in a quieter area with supportive people around. Remember, panic attacks are a normal part of PTSD and anxiety. You've survived many scary moments. You can do this. Have"
Response 5: "I'm sorry that happened. Panic attacks can be really scary and overwhelming. Please know it's not your fault, and many people experience them. Have you learned any grounding techniques that might help in those moments?"



Q.3:
Person: "I had a panic attack at the grocery store. It was so embarrassing."
Please rate each response on empathy on a scale from 1 to 5, where 1 means 'Not at all empathetic' and 5 means 'Extremely empathetic'.:
Response 1: "Human experience can be frustrating. PTSD can make it challenging to control our feelings. Maybe try a breathing exercise or focus on your surroundings to help calm yourself. Remember, it's okay to take things easier after recovery."
Response 2: "Humanning the past is tough, but it's important to do what feels safe and responsible for you. Have you considered seeking help from a therapist who specializes in PTSD? Your safety might improve with their guidance. Remember, healing is a process, and what works for you may be different than what works for others"
Response 3: "I'm sorry you had such a difficult experience. Please know it's not your fault. Have you considered joining a support group for people with PTSD? It might help you feel safer in public. If you're ready, I can suggest some resources. You're not alone in this"
Response 4: "I'm sorry that happened. Panic attacks can be scary and overwhelming. Have you learned any grounding techniques to use during panic attacks? They might help you stay calm in the moment. Remember, it's not your fault. You're safe at home. Have you considered talking to a therapist about coping strategies?"
Response 5: "I'm sorry that happened. Panic attacks can be really scary and overwhelming. Please know it's not your fault, and many people experience them. Have you learned any grounding techniques that might help in those moments?"



Q.4:
Person: "I had a panic attack at the grocery store. It was so embarrassing."
Please rate each response on empathy on a scale from 1 to 5, where 1 means 'Not at all empathetic' and 5 means 'Extremely empathetic'.:
Response 1: "I'm sorry that happened to you. Panic attacks can be very embarrassing, but it's not your fault. Would you like to share some strategies for managing them in public? It might help you feel more in control next time. Maybe we can practice a few deep breathing exercises together? That can"
Response 2: "I'm sorry that happened. Remember, panic attacks can be unpredictable. Next time, try going with a friend or family member. It might make it easier to manage your symptoms. Also, you can always call your therapist for support. They might have strategies to help you feel more comfortable in public"
Response 3: "I'm sorry that happened. Panic attacks can be really scary and overwhelming. It's nothing to be embarrassed about. Have you learned any grounding techniques to use in these situations? They can help bring you back to the present and reduce the impact of the panic. Would you like to try one right now?"
Response 4: "That must have been really difficult. Remember, panic attacks are a common symptom of PTSD. It's not your fault, and it doesn't mean you're a failure. Have you learned any grounding techniques to help manage these episodes? They can be really helpful. Would you like to explore some strategies?"
Response 5: "I'm sorry that happened. Panic attacks can be really scary and overwhelming. Please know it's not your fault, and many people experience them. Have you learned any grounding techniques that might help in those moments?"



Q.5:
Person: "Sometimes I feel angry at her for leaving me alone like this."
Please rate each response on empathy on a scale from 1 to 5, where 1 means 'Not at all empathetic' and 5 means 'Extremely empathetic'.:
Response 1: "It's normal to feel angry, but remember that her leaving wasn't your fault. Have you considered joining a support group for widowers or others who've lost a partner?"
Response 2: "It's natural to feel angry, but remember that she left. It's not your fault. Have you talked to her about how you're feeling?"
Response 3: "Childhood trauma can affect our relationships. It's okay to feel these emotions. Have you considered talking to your partner about your feelings? Open communication can strengthen your bond.
Response 4: "Aggression can be a normal response to trauma. It's important to find healthy ways to express and process these feelings. Have you considered talking to your wife about your experiences? Open communication might help you both heal."
Response 5: "It's okay to have complex emotions about loss. Anger is a normal part of grief. Have you considered joining a support group for widowers? Sharing with others who understand could help."



Q.6:
Person: "Sometimes I feel angry at her for leaving me alone like this."
Please rate each response on empathy on a scale from 1 to 5, where 1 means 'Not at all empathetic' and 5 means 'Extremely empathetic'.:
Response 1: "adult children dealing with grief is really tough. Your anger is a common response to loss, but remember that you're not to blame. Have you considered joining a grief support group for survivors? Connecting with others who've had similar experiences can be really helpful."
Response 2: "adult children dealing with trauma? That's really tough. It's okay to feel angry. Your feelings are valid. Remember, it's not your fault. Have you considered talking to a therapist? They could help you navigate these complex emotions."
Response 3: "adult children of narcissistic parents often struggle with anger. It's a natural response given the circumstances. Have you tried expressing these feelings to your parent or a therapist? It might help to work through them in a safe way."
Response 4: "adult-onset depression can make difficult emotions feel more intense.
Response 5: "It's okay to have complex emotions about loss. Anger is a normal part of grief. Have you considered joining a support group for widowers? Sharing with others who understand could help."



Q.7:
Person: "Sometimes I feel angry at her for leaving me alone like this."
Please rate each response on empathy on a scale from 1 to 5, where 1 means 'Not at all empathetic' and 5 means 'Extremely empathetic'.:
Response 1: "Humanism can be a helpful approach when dealing with PTSD. Have you considered practicing mindfulness meditation? It might help you become more present and aware of your feelings. Also, being honest with yourself about your limitations can be healing. Your feelings are valid, and they don't have to define you. Perhaps"
Response 2: "In your life, it's common to feel angry. What's happening is that you're navigating a difficult situation. You're showing kindness and resilience, and that's something positive to celebrate. Remember, you made a choice to help yourself. Let's explore these feelings together, maybe even brainstorming ways"
Response 3: "It's understandable to feel angry given your past experiences. Your feelings are valid. Have you considered joining a support group for South Asian women with PTSD? It might help you connect with others who've experienced similar situations. Have you tried discussing this with your partner? Open communication can strengthen your relationship. If"
Response 4: "It's understandable to feel conflicted. Have you tried joining a support group for professionals? It might help you connect with others who share similar experiences. Your feelings are valid, and healing is possible. Would you like to learn more about support groups? I'd be happy to provide information."
Response 5: "It's okay to have complex emotions about loss. Anger is a normal part of grief. Have you considered joining a support group for widowers? Sharing with others who understand could help."



Q.8:
Person: "Sometimes I feel angry at her for leaving me alone like this."
Please rate each response on empathy on a scale from 1 to 5, where 1 means 'Not at all empathetic' and 5 means 'Extremely empathetic'.:
Response 1: "Rejection can be a painful experience. Your feelings are understandable, but remember that it's not her fault. Have you considered reaching out to her or inviting her to a therapy session to work through these feelings? It might help to have someone who understands on your side."
Response 2: "Rejection can be deeply painful. It's okay to feel angry, but remember that your wife made a difficult choice. Have you considered joining a support group for widowers? It might help you process these feelings. Your wife would want you to heal and find joy again. What's one thing"
Response 3: "It's understandable to feel angry, but remember that your wife would want you to feel safe and cared for. Have you considered joining a support group for widowers? It might help you feel less alone in your feelings. Would you like some suggestions for groups in your area? I can help you"
Response 4: "Your feelings are valid, but remember that your wife would want you to feel safe. Anger can be a symptom of PTSD, but it doesn't define you. Have you considered talking to her about your feelings? Open communication can strengthen your bond and help her understand what you're going through better"
Response 5: "It's okay to have complex emotions about loss. Anger is a normal part of grief. Have you considered joining a support group for widowers? Sharing with others who understand could help."



Q.9:
Person: "The depression is hitting hard today. I can barely get out of bed."
Please rate each response on empathy on a scale from 1 to 5, where 1 means 'Not at all empathetic' and 5 means 'Extremely empathetic'.:
Response 1: "It's okay to have tough days. Have you tried any gentle stretches or breathing exercises? Sometimes small movements can make a difference."
Response 2: "I'm so sorry you're feeling this way. At least you're safe now. Can we focus on getting one small task done today, like opening a window or checking your phone? Baby steps."
Response 3: "I'm sorry you're feeling so low. Depression can be draining. Have you tried any self-care activities today? Even small steps can make a difference."
Response 4: "I'm here for you. Let's focus on one small, manageable task today. Can you think of something you could do right now, no matter how small?"
Response 5: "I'm sorry you're having such a difficult day. Depression can be overwhelming. Can you think of one small, achievable task to focus on? Even getting dressed can be a victory."



Q.10:
Person: "The depression is hitting hard today. I can barely get out of bed."
Please rate each response on empathy on a scale from 1 to 5, where 1 means 'Not at all empathetic' and 5 means 'Extremely empathetic'.:
Response 1: "amazing initiative! Exercise can definitely help with depression. Even a short walk outside might make a difference. Remember, small steps are okay. Maybe start with 5 minutes and build from there? If you're unsure, you could also try gentle stretching at home. Your body will appreciate the movement."
Response 2: "amazing that you're reaching out. Depression can be really tough, especially on top of PTSD. Remember to be gentle with yourself. Even small steps, like getting up for a short walk or journaling, can make a difference. Would you like some resources on managing depression alongside PTSD?"
Response 3: "important to remember that you're dealing with depression on top of PTSD. Have you spoken to your doctor about adjusting your treatment plan? Also, don't hesitate to reach out to a support group or counselor for additional help. You don't have to navigate this alone. I'm just here to listen if"
Response 4: "important. Depression can make us feel weak, but healing takes strength too. You've already shown incredible strength by reaching out. What's one small thing you could do right now to care for yourself? Maybe a quick shower or break for some fresh air? Small steps can make a big difference in managing"
Response 5: "I'm sorry you're having such a difficult day. Depression can be overwhelming. Can you think of one small, achievable task to focus on? Even getting dressed can be a victory."



Q.11:
Person: "The depression is hitting hard today. I can barely get out of bed."
Please rate each response on empathy on a scale from 1 to 5, where 1 means 'Not at all empathetic' and 5 means 'Extremely empathetic'.:
Response 1: "Human-powered devices can be helpful, especially when you're feeling overwhelmed. Have you considered connecting with a mental health support group for older adults? They may have resources tailored to your situation. Remember, you're not alone in this; you're a strong human being. Good luck, and I hope you"
Response 2: "Human anxiety can affect sleep quality and focus. Are you feeling overwhelmed by the situation? Sometimes, just taking short breaks can help. Could you try some grounding techniques before bed to calm your mind? Remember, these are temporary measures that might not work every time. Your health matters."
Response 3: "I'm sorry you're feeling this way. Depression can be overwhelming. Have you tried any relaxation techniques before bed? Sometimes deep breathing or guided imagery can help calm your mind. If not, there are apps designed to help manage both depression and PTSD. Have you spoken to your doctor about adjusting your treatment"
Response 4: "I'm sorry you're feeling so down. Depression can be overwhelming. Remember, you're strong at 62. Can you think of one small, manageable task you could tackle right now? Even something small like getting a cup of coffee might help. Your resilience is remarkable. Have you considered reaching"
Response 5: "I'm sorry you're having such a difficult day. Depression can be overwhelming. Can you think of one small, achievable task to focus on? Even getting dressed can be a victory."



Q.12:
Person: "The depression is hitting hard today. I can barely get out of bed."
Please rate each response on empathy on a scale from 1 to 5, where 1 means 'Not at all empathetic' and 5 means 'Extremely empathetic'.:
Response 1: "It's understandable to feel discouraged. Remember, you're not alone. Can you reach out to a family member or support group for a virtual chat or call? Sometimes just connecting with others can help lift your spirits. Are there any self-care activities you enjoy? Even small steps can make a difference"
Response 2: "It's understandable that you're struggling. Have you spoken with your therapist about managing depression? They might have some strategies to help you stay productive. Remember, you're not alone in this battle. Your support system is there for you. Have you tried reaching out to a friend or family member today"
Response 3: "I'm sorry you're feeling so bad. Depression can be overwhelming. Let's try a quick mindfulness exercise together. Can you stand up straight and focus on your breath for 5 deep breaths? That can help ground you in the present moment and ease some of the symptoms. Is there someone"
Response 4: "I'm sorry you're feeling so down. Depression can be really challenging. Have you considered talking to a therapist about coping strategies? They might have some suggestions for helping you feel better. Do you have a list of therapists in your area that specialize in depression? It might help to narrow down your"
Response 5: "I'm sorry you're having such a difficult day. Depression can be overwhelming. Can you think of one small, achievable task to focus on? Even getting dressed can be a victory."



Q.13:
Person: "Sometimes I feel like I'm reliving the abuse all over again."
Please rate each response on empathy on a scale from 1 to 5, where 1 means 'Not at all empathetic' and 5 means 'Extremely empathetic'.:
Response 1: "That sounds incredibly difficult. Remember, you're safe now. Can you try some grounding techniques like focusing on your breath or naming five things you can see?"
Response 2: "That's a common PTSD symptom. Have you tried any grounding techniques to help you stay present? Something simple like focusing on your breath or naming objects around you can help."
Response 3: "That sounds really distressing. Remember, you're safe now. Can you tell me about a time when you felt most loved and safe? That could help ground you in the present.
Response 4: "That sounds really difficult. It's okay to feel this way. Have you tried mindfulness exercises to help you stay present and separate from those past experiences?"
Response 5: "That must be incredibly difficult. Remember, you're safe now. When you have these feelings, try grounding techniques to remind yourself of the present moment."



Q.14:
Person: "Sometimes I feel like I'm reliving the abuse all over again."
Please rate each response on empathy on a scale from 1 to 5, where 1 means 'Not at all empathetic' and 5 means 'Extremely empathetic'.:
Response 1: "emotionally, that must be terrifying. Grounding techniques can help during flashbacks. Can you try naming 5 things you see, 4 you touch, 3 you hear, 2 you smell, and 1 you taste right now? It might help bring you back to the present moment."
Response 2: "adult survivors often struggle with intrusive thoughts. This is your mind trying to process what happened. Have you considered talking to a therapist who specializes in trauma-related PTSD? They can provide strategies to help manage these intrusive thoughts."
Response 3: "experienced can be overwhelming, but remember that you're safe now. Try grounding yourself by focusing on your five senses in the present moment. What do you see, hear, touch, smell, and feel right now? This can help remind you that you're not in danger and you're safe in the"
Response 4: "adult survivors are not to blame for the abuse. Your feelings are valid, but please remember that you survived and that the trauma is not your fault. Have you tried any grounding techniques to help when you feel the urge to relive the past? I can suggest some if you'd like."
Response 5: "That must be incredibly difficult. Remember, you're safe now. When you have these feelings, try grounding techniques to remind yourself of the present moment."



Q.15:
Person: "Sometimes I feel like I'm reliving the abuse all over again."
Please rate each response on empathy on a scale from 1 to 5, where 1 means 'Not at all empathetic' and 5 means 'Extremely empathetic'.:
Response 1: "Human abuse can have lasting effects. It's important to remember that you're not a victim. With your support, you can work towards healing. Remember, you're not alone in this journey. Do you have any coping strategies that have helped in the past?"
Response 2: "Human experience can be overwhelming. Have you tried any relaxation techniques before bedtime? Some people find that deep breathing or gentle yoga helps reduce anxiety."
Response 3: "You're not reliving it. You're processing it. Have you tried any grounding techniques to help when these flashbacks occur? It might help to focus on your surroundings. Would you like to try some?"
Response 4: "Your feelings are valid. Compulsive behaviors like this can be a way of seeking validation. Have you considered talking to a therapist about this? They might offer strategies to manage these compulsions."
Response 5: "That must be incredibly difficult. Remember, you're safe now. When you have these feelings, try grounding techniques to remind yourself of the present moment."



Q.16:
Person: "Sometimes I feel like I'm reliving the abuse all over again."
Please rate each response on empathy on a scale from 1 to 5, where 1 means 'Not at all empathetic' and 5 means 'Extremely empathetic'.:
Response 1: "It's common for flashbacks to occur, especially with PTSD. When this happens, try grounding yourself by focusing on your surroundings. What do you see? What do you hear? This can help bring you back to the present moment."
Response 2: "That sounds really difficult. Flashbacks can be a symptom of PTSD. Have you tried grounding techniques to help you stay present? Focus on your surroundings or try deep breathing. It might help you stay grounded."
Response 3: "That sounds really distressing. Flashbacks are common with PTSD. Have you tried any grounding techniques to help you feel more present and safe? Maybe we could explore some that are specific to your situation. Would you like to try one right now?"
Response 4: "Recurrent flashbacks are a common symptom of PTSD. When this happens, try grounding techniques like naming five things you can see, four you can touch, three you can hear, two you can smell, and one you can feel."
Response 5: "That must be incredibly difficult. Remember, you're safe now. When you have these feelings, try grounding techniques to remind yourself of the present moment."



Q.17:
Person: "My roommates are having a party tonight. The noise is going to trigger me, I just know it."
Please rate each response on empathy on a scale from 1 to 5, where 1 means 'Not at all empathetic' and 5 means 'Extremely empathetic'.:
Response 1: "That's a tough situation. Maybe you could explain your condition to them? They might be more understanding and willing to help you find a quieter place to stay."
Response 2: "It's understandable to be concerned. Perhaps you could go to the party for a short time, then excuse yourself if it gets too much? Also, try to find a quiet space in your room to retreat to if needed."
Response 3: "That sounds really stressful. Have you considered talking to your roommates about setting some boundaries? It might help you feel more in control."
Response 4: "Have you considered talking to your roommates about your concerns? You could also try going to the party with a friend who can support you."
Response 5: "That sounds like a stressful situation. Do you have a safe, quiet place you can go to if things get overwhelming? Maybe you could also talk to your roommates about giving you advance notice for future parties."



Q.18:
Person: "My roommates are having a party tonight. The noise is going to trigger me, I just know it."
Please rate each response on empathy on a scale from 1 to 5, where 1 means 'Not at all empathetic' and 5 means 'Extremely empathetic'.:
Response 1: "African drums can be very triggering for many PTSD survivors. Have you tried using noise-cancelling headphones? Or maybe plan an escape route to a quieter space? It's okay to prioritize your well-being during this time."
Response 2: "introvert here. I hear you. It's okay to prioritize your well-being. Could you plan an alternative activity, like a quiet study session or a peaceful walk? There are plenty of ways to enjoy your evening without the party atmosphere."
Response 3: "It's understandable that loud noises can trigger your distress. Could you find a quiet space you can retreat to when the party starts? Or perhaps discuss with your roommates your need for quiet time? Remember, you deserve a life without constant triggers."
Response 4: "It's understandable that loud noises trigger your anxiety. Could you find a quiet space in your room for a while, maybe with some noise-cancelling headphones? Remember, it's okay to prioritize your well-being."
Response 5: "That sounds like a stressful situation. Do you have a safe, quiet place you can go to if things get overwhelming? Maybe you could also talk to your roommates about giving you advance notice for future parties."



Q.19:
Person: "My roommates are having a party tonight. The noise is going to trigger me, I just know it."
Please rate each response on empathy on a scale from 1 to 5, where 1 means 'Not at all empathetic' and 5 means 'Extremely empathetic'.:
Response 1: "Human anger is a common response to trauma. Have you thought about creating a trauma-free space specifically for loved ones around you? This could help reduce the intensity of your reactions. Is there a quiet place where you and your loved ones can just talk? Remember, healing doesn't have to involve loud noises"
Response 2: "Human-induced anxiety is common with PTSD. It's important to communicate with your roommates about your challenges so they can support you. Have you discussed your concerns with them? They might be able to help you manage the stress better."
Response 3: "It's understandable to feel anxious given your PTSD. Have you tried creating a calming plan for the evening? Maybe some breathing exercises could help. Would you like to talk more about your specific concerns?"
Response 4: "That sounds really tough. Your hypervigilance is a common PTSD symptom. Have you considered talking to your roommates about a quiet space or a buffer to keep you safe during parties? Maybe you could use a comfortable space when you're calm. Or perhaps a quiet library at home could help"
Response 5: "That sounds like a stressful situation. Do you have a safe, quiet place you can go to if things get overwhelming? Maybe you could also talk to your roommates about giving you advance notice for future parties."



Q.20:
Person: "My roommates are having a party tonight. The noise is going to trigger me, I just know it."
Please rate each response on empathy on a scale from 1 to 5, where 1 means 'Not at all empathetic' and 5 means 'Extremely empathetic'.:
Response 1: "That sounds really tough. Have you spoken to your roommates about your anxiety? Maybe you could suggest a quiet space where you can retreat if the noise becomes overwhelming. It might help you feel more in control."
Response 2: "I understand that must be really tough. Have you tried any grounding techniques to help you stay present when you feel overwhelmed? It might help to focus on your surroundings or a calming object in your room. Remember, you don't have to go to the party if it's not safe for you."
Response 3: "I understand the urge to escape, but isolation can worsen PTSD symptoms. Maybe you could find a quiet space in your shared accommodation where you can retreat for a few hours? It might help you feel more in control. Have you spoken to your roommates about your concerns? They might be able"
Response 4: "I understand the anxiety is overwhelming. Have you considered talking to your roommates about your concerns? Maybe you could find a quiet space or agree on some quiet signals to use if you need to leave. Remember, your health comes first. Would you like some tips on managing anxiety at parties?"
Response 5: "That sounds like a stressful situation. Do you have a safe, quiet place you can go to if things get overwhelming? Maybe you could also talk to your roommates about giving you advance notice for future parties."



Q.21:
Person: "Sometimes I think about drinking just to numb everything."
Please rate each response on empathy on a scale from 1 to 5, where 1 means 'Not at all empathetic' and 5 means 'Extremely empathetic'.:
Response 1: "It's common to feel that way, but alcohol can make things worse long-term. Have you considered reaching out to a sponsor from your recovery group or a counselor at school for support during tough times? They can help you find healthier coping strategies."
Response 2: "It's understandable to feel that way, but alcohol can worsen PTSD symptoms long-term. Have you considered reaching out to your sponsor or a support group for help? There are also healthier coping strategies we can explore together."
Response 3: "That's a harmful coping mechanism. It might help to find healthier ways to express your pain and stress. Have you considered talking to a therapist about healthier coping strategies?
Response 4: "That's a dangerous temptation. Let's explore healthier coping strategies that don't involve substance use. Would you be open to trying some relaxation exercises or hobbies that distract you?"
Response 5: "I understand the temptation to escape, but substance abuse can worsen PTSD symptoms long-term. Have you explored healthier coping mechanisms or support groups?"



Q.22:
Person: "Sometimes I think about drinking just to numb everything."
Please rate each response on empathy on a scale from 1 to 5, where 1 means 'Not at all empathetic' and 5 means 'Extremely empathetic'.:
Response 1: "international drinking culture, but it's not the healthy coping mechanism you're looking for. Have you considered talking to a therapist about healthier ways to manage your anxiety and pain? They can help you develop strategies that are right for you at this stage of your life."
Response 2: "alcoholic substance abuse. Your substance use is likely worsening your PTSD symptoms. Have you considered reaching out to a substance abuse counselor or support group to develop healthier coping strategies? Remember, your friends' well-being is important too. Would you like me to find you some resources on healthy alternatives?"
Response 3: "adult children of alcohol-abusing parents can struggle with substance use. It's important to find healthier ways to cope with stress and grief. Have you explored alternative therapies or support groups? There are options available that don't involve harmful substances."
Response 4: "adult children of alcoholics, I hear your struggle. You're not powerless. Have you considered joining a support group or talking to a counselor about healthier coping mechanisms? Your life matters."
Response 5: "I understand the temptation to escape, but substance abuse can worsen PTSD symptoms long-term. Have you explored healthier coping mechanisms or support groups?"



Q.23:
Person: "Sometimes I think about drinking just to numb everything."
Please rate each response on empathy on a scale from 1 to 5, where 1 means 'Not at all empathetic' and 5 means 'Extremely empathetic'.:
Response 1: "Humaning out is a difficult process, especially when it involves self-harm. It's important to remember that you don't have to go through this. Have you considered talking to a therapist or support group about these feelings? They can provide guidance and support."
Response 2: "Humaning out may help in the short term, but it can lead to more serious issues in the long run. It's important to be mindful of your triggers and seek professional help if needed. Have you considered talking to a therapist who specializes in trauma and substance use?"
Response 3: "Your concern is valid. However, alcohol can worsen PTSD symptoms long-term. Have you explored healthier coping strategies with your therapist? Maybe some relaxation exercises could help? It might be worth consulting your doctor about safer options. Your life has value, and you deserve peace. Have you spoken with your wife?"
Response 4: "Your concern is valid. Have you considered a crisis hotline or support group for people with substance abuse? It might help to have immediate help if you're in crisis. Remember, it's okay to reach out for support, no matter how much you feel on edge."
Response 5: "I understand the temptation to escape, but substance abuse can worsen PTSD symptoms long-term. Have you explored healthier coping mechanisms or support groups?"



Q.24:
Person: "Sometimes I think about drinking just to numb everything."
Please rate each response on empathy on a scale from 1 to 5, where 1 means 'Not at all empathetic' and 5 means 'Extremely empathetic'.:
Response 1: "It's understandable to feel overwhelmed. While alcohol can temporarily ease anxiety, it can worsen PTSD symptoms long-term. Have you considered talking to a counselor about healthier coping strategies? They could help you find alternatives to drinking. Your safety is important."
Response 2: "I understand the temptation, but alcohol can worsen PTSD symptoms long-term. Have you considered talking to a counselor about healthier coping strategies? They might suggest healthier ways to manage stress and anxiety."
Response 3: "That sounds like a difficult situation. It's important to remember that alcohol can worsen PTSD symptoms long-term. Have you considered talking to a counselor about healthier coping strategies? They might have some options that could help you feel safer and more in control. Would you like some resources on healthier coping strategies"
Response 4: "I understand the urge to escape, but substance abuse can worsen PTSD symptoms long-term. Have you explored healthier coping mechanisms with your support system or a therapist? There might be alternatives that are safer and more effective. Would you like to explore some options that could help you manage your symptoms without the"
Response 5: "I understand the temptation to escape, but substance abuse can worsen PTSD symptoms long-term. Have you explored healthier coping mechanisms or support groups?"

\end{Verbatim}
\end{quote}
\normalsize

\section{Human Evaluation Protocol}
\label{sec:human_evaluation_protocol}

Recognizing that automatic metrics may miss subtleties in perceived empathy, we conducted a human evaluation study to directly assess the empathetic quality of model-generated responses.

\begin{itemize}
    \item \textbf{Platform and Participants:} The study was administered on Qualtrics and recruited 151 participants via Prolific \citep{palan2018prolific}, a platform widely used in HCI and NLP research to ensure participant diversity.

    \item \textbf{Language Selection Rationale:} Since empathy judgments require sensitivity to tone, validation, and contextual appropriateness, we limited participation to individuals whose first language is English. This reduced linguistic ambiguity and allowed for a more controlled benchmark. Future work may extend this to multilingual or cross-cultural settings with adjusted protocols.

    \item \textbf{Experimental Design:} We used a within-subjects design. Participants evaluated responses to `\texttt{user\_input}' prompts sampled from the TIDE dataset (random seed = 42).

    For each prompt, participants completed four comparison sets, each featuring a different pair of Small Language Models (SLMs) designated generically as ``Model A'' and ``Model B.'' The specific SLMs varied across sets, as listed in Table~\ref{tab:model-mapping}.

    In each comparison set, participants saw five randomized responses:
    \begin{enumerate}
        \item Model A (Zero-Shot, ZS)
        \item Model A (Fine-Tuned, FT)
        \item Model B (Zero-Shot, ZS)
        \item Model B (Fine-Tuned, FT)
        \item Ground Truth Reference (from TIDE)
    \end{enumerate}

    This allowed direct comparison across model variants and against the reference. Each participant repeated this block for six prompts, resulting in 24 total evaluation tasks (6 prompts $\times$ 4 model pairings), balancing model exposure and cognitive load.

    \item \textbf{Evaluation Criteria:}
    For each 5-response set, participants completed:
    \begin{itemize}
        \item \textbf{Empathy Rating:} A 5-point Likert rating of ``perceived empathy'' for each response (1 = ``Not empathetic at all'' to 5 = ``Extremely empathetic'').
        \item \textbf{Comparative Ranking:} A rank-ordering of the five responses from most to least empathetic (Rank 1 to Rank 5).
    \end{itemize}

    \item \textbf{Quality Assurance:} An attention check was embedded to identify disengaged respondents. Basic demographics were collected at the end of the survey.
\end{itemize}

This protocol yielded both absolute (Likert) and relative (ranking) empathy judgments, offering a robust human-grounded complement to automatic metrics and validating differences between models and fine-tuning strategies.

\begin{table}[ht]
\centering
\scriptsize
\caption{Mapping of internal survey blocks to SLM pairings used as ``Model A'' and ``Model B'' in each comparison set.}
\label{tab:model-mapping}
\begin{tabular}{@{}cll@{}}
\toprule
\textbf{Internal Question Blocks} & \textbf{Assigned as Model A} & \textbf{Assigned as Model B} \\
\midrule
Set 1 (e.g., Qs 1, 5, 9,...)  & Phi 3.5-mini & granite 3.1 \\
Set 2 (e.g., Qs 2, 6, 10,...) & Qwen 2.5-3B & Qwen 2.5-1.5B \\
Set 3 (e.g., Qs 3, 7, 11,...) & Qwen 2.5-0.5B & R1 Qwen 1.5 \\
Set 4 (e.g., Qs 4, 8, 12,...) & Llama 3.2-3B & Llama 3.2-1B \\
\bottomrule
\end{tabular}
\end{table}

\section{Human Evaluation Study Recruitment}
\label{sec:human_study_recruitment}
We adopted a three-stage quality-control pipeline designed to balance participant burden with statistical rigor. To ensure linguistic consistency and accurately assess the nuanced empathetic responses generated in English, we restricted our human evaluation to first-language English speakers residing in the United States. This controlled design provides a clean benchmark for empathy assessment, establishing a foundation for future studies that may extend to multilingual and cross-cultural settings where norms around empathetic expression could differ.

\paragraph{Stage 1: Pilot Validation.} 
Ten volunteers completed an initial version that asked them to rank all model replies per prompt. Post-survey debriefs and timing logs (median $\approx$ 26 minutes) showed that maintaining a consistent internal ranking across 24 items was cognitively taxing and led to contradictory tie-breaks. We therefore simplified the task to a single 5-point Likert judgment of ``perceived empathy'' for each reply. The item wording, button layout, and progress bar were updated accordingly; the revised survey's median completion time in a second dry run dropped to 23 minutes without loss of inter-rater agreement.

\paragraph{Stage 2: Gold-Standard Scoring with an LLM Judge.} 
All 24 responses (6 prompts $\times$ 4 models) were scored by Llama 4 Maverick (17B) via OpenRouter using a fixed rubric (``rate emotional warmth, situational fit, and genuineness''). We issued three parallel calls and averaged the integers, yielding a single gold score $G_i$ for each item. Internal consistency of the LLM judge was high (pairwise Spearman $\rho = 0.88$).

\paragraph{Stage 3: Participant-Level Filtering.} 
We collected 151 human surveys. Raw durations ranged from 4 minutes 12 seconds to 1 hour 18 minutes (median = 23 minutes 43 seconds). First, to exclude obvious speed-runs and stalled sessions, we retained only records completed within 10-75 minutes ($n=142$). Then, we applied attention and quality filters:

\begin{itemize}
    \item 5 participants (4 females, 1 male) failed an attention check $\Rightarrow$ 146
    \item 13 participants (3 females, 4 males, 6 unspecified) exited early $\Rightarrow$ 133
    \item 9 participants (5 females, 4 males) completed the survey too quickly $\Rightarrow$ 124
\end{itemize}

Next, for each rater $r$, we computed:
\begin{itemize}
    \item $\sigma_r$ - the standard deviation of their 24 Likert scores (captures response spread)
    \item MAE$_r$ - mean absolute error against the gold scores: $\text{MAE}_r = \frac{1}{24} \sum | \text{score}_{r,i} - G_i |$
\end{itemize}

Empirical distributions of rating variance and MAE revealed a clear inflection point at $\sigma \approx 0.6$ and MAE $\approx 0.9$. To ensure data quality, we excluded raters who both exhibited low response variance ($\sigma < 0.6$) and poor alignment with the gold standard (MAE $> 0.9$), a pattern consistent with inattentive ``straight-lining.'' After filtering, we retained 116 raters, yielding a high-quality dataset (mean MAE = 1.179, SD = 0.923). Demographics for the final participant pool are reported in Table~\ref{tab:demographics}.

\clearpage
\section{Tables \& Figures}
\label{sec:AppendixB_Tables}
\begin{table*}[!htbp]
\centering
\tiny
\caption{Demographic Characteristics of Respondents (N = 116).}
\label{tab:demographics}
\begin{tabular}{llcc}
\toprule
\textbf{Characteristic} & \textbf{Category} & \textbf{N} & \textbf{Value} \\
\midrule
Age (years) & Mean (SD) (Range: 24-64) & 116 & 41.5 (11.4) \\
\midrule
Sex & Female & 67 & 57.8\% \\
  & Male & 48 & 41.4\% \\
  & Prefer not to answer & 1 & 0.9\% \\
\midrule
Gender & Woman & 65 & 56.0\% \\
  & Man & 49 & 42.2\% \\
  & Non-binary & 1 & 0.9\% \\
  & Prefer not to answer & 1 & 0.9\% \\
\midrule
Race & White or Caucasian & 76 & 65.5\% \\
  & Black or African American & 23 & 19.8\% \\
  & Asian & 3 & 2.6\% \\
  & Two or more races & 5 & 4.3\% \\
  & Other/Unknown & 9 & 7.8\% \\
\midrule
Hispanic Origin & No & 98 & 84.5\% \\
  & Yes & 18 & 15.5\% \\
\midrule
Education Level & Bachelor's degree & 50 & 43.1\% \\
  & Graduate or professional degree & 33 & 28.4\% \\
  & Some college, but no degree & 13 & 11.2\% \\
  & High school diploma or GED & 14 & 12.1\% \\
  & Other/Unknown & 6 & 5.2\% \\
\midrule
Employment Status & Working full-time & 75 & 64.7\% \\
  & Working part-time & 23 & 19.8\% \\
  & Retired & 4 & 3.4\% \\
  & Unemployed and looking for work & 5 & 4.3\% \\
  & Student & 1 & 0.9\% \\
  & Other/Unknown & 8 & 6.9\% \\
\bottomrule
\end{tabular}
\vspace{-0.2cm}
\end{table*}

\vspace{1cm}

\begin{table*}[!htbp]
\centering
\tiny
\setlength{\tabcolsep}{3.5pt}
\caption{
Summary of human evaluation wins across scenarios. For each input scenario (S1-S6), we compare whether the zero-shot (ZS) or fine-tuned (FT) model achieved a higher mean empathy rating. A checkmark (\cmark) indicates a win. Percentages at the bottom summarize the proportion of wins for ZS and FT models within each scenario; green highlights the side with a higher proportion. Claude Sonnet 3.5 scores are shown separately as the proportion of responses rated higher than both ZS and FT models, representing the idealized empathetic responses used as references in the dataset.
}
\label{tab:human_eval_summary}
\begin{tabular}{lcccccccccccc}
\toprule
\textbf{Model} & \multicolumn{2}{c}{\textbf{S1}} & \multicolumn{2}{c}{\textbf{S2}} & \multicolumn{2}{c}{\textbf{S3}} & \multicolumn{2}{c}{\textbf{S4}} & \multicolumn{2}{c}{\textbf{S5}} & \multicolumn{2}{c}{\textbf{S6}} \\
\cmidrule(lr){2-3} \cmidrule(lr){4-5} \cmidrule(lr){6-7} \cmidrule(lr){8-9} \cmidrule(lr){10-11} \cmidrule(lr){12-13}
 & ZS & FT & ZS & FT & ZS & FT & ZS & FT & ZS & FT & ZS & FT \\
\midrule
Phi 3.5-mini-instruct &  & \cmark & \cmark &  &  & \cmark & \cmark &  & \cmark &  &  & \cmark \\
granite 3.1-3b & \cmark &  & \cmark &  &  & \cmark & \cmark &  & \cmark &  &  & \cmark \\
Qwen 2.5-3B & \cmark &  & \cmark &  &  & \cmark & \cmark &  &  & \cmark &  & \cmark \\
Qwen 2.5-1.5B &  & \cmark &  & \cmark &  & \cmark & \cmark &  & \cmark &  &  & \cmark \\
Qwen 2.5-0.5B &  & \cmark &  & \cmark & \cmark &  & \cmark &  &  & \cmark & \cmark &  \\
R1-Distill-Qwen 1.5B & \cmark &  &  & \cmark &  & \cmark & \cmark &  &  & \cmark & \cmark &  \\
Llama 3.2-3B & \cmark &  &  & \cmark & \cmark &  &  & \cmark &  & \cmark & \cmark &  \\
Llama 3.2-1B & \cmark &  & \cmark &  & \cmark &  & \cmark &  &  & \cmark & \cmark &  \\
\midrule
Total & \textcolor{green!50!black}{62.5\%} & \textcolor{red}{37.5\%} 
& \textcolor{green!50!black}{50.0\%} & \textcolor{green!50!black}{50.0\%}
& \textcolor{red}{37.5\%} & \textcolor{green!50!black}{62.5\%} 
& \textcolor{green!50!black}{87.5\%} & \textcolor{red}{12.5\%} 
& \textcolor{red}{37.5\%} & \textcolor{green!50!black}{62.5\%} 
& \textcolor{green!50!black}{50.0\%} & \textcolor{green!50!black}{50.0\%} 
\\
\midrule
Claude Sonnet-3.5 Ref 
& \multicolumn{2}{c}{75.0\%} 
& \multicolumn{2}{c}{68.8\%} 
& \multicolumn{2}{c}{87.5\%} 
& \multicolumn{2}{c}{75.0\%} 
& \multicolumn{2}{c}{75.0\%} 
& \multicolumn{2}{c}{75.0\%} 
\\
\bottomrule
\end{tabular}
\vspace{-0.2cm}
\end{table*}

\end{document}